\documentclass[10pt,twocolumn,letterpaper]{article}

\usepackage[margin=1in]{geometry}
\usepackage{amsmath,amssymb}
\usepackage{newtxtext,newtxmath}
\usepackage{graphicx}
\usepackage{booktabs}
\usepackage{multirow}
\usepackage[numbers,sort&compress]{natbib}
\usepackage{xcolor}
\definecolor{cvprblue}{rgb}{0.21, 0.49, 0.74}
\usepackage[colorlinks=true,allcolors=cvprblue]{hyperref}

\title{
CLIP-RD: Relational Distillation for Efficient CLIP Knowledge Distillation
}

\author{
Jeannie Chung$^{*}$,
Hanna Jang$^{*}$,
Ingyeong Yang$^{*}$,
Uiwon Hwang$^{\dagger}$,
Jaehyeong Sim$^{\dagger}$ \\
Ewha Womans University, Seoul 03760, South Korea \\
\texttt{uiwon.hwang@ewha.ac.kr, jh.sim@ewha.ac.kr} \\
$^{*}$ Equal contribution \quad
$^{\dagger}$ Corresponding authors
}
\date{}

\begin{document}

\maketitle
\begin{abstract}
Contrastive Language-Image Pre-training (CLIP) demonstrates strong zero-shot generalization, but due to substantial computational and memory costs, distillation into lightweight models is required. Existing relational objectives do not explicitly model multidirectional relationships between teacher and student embeddings, potentially leaving the geometric relationships insufficiently constrained. This may disrupt the modality-gap structure important for zero-shot transfer.
To address these limitations, we propose a relational distillation framework, \textbf{CLIP-RD}, which introduces two relational methods, Vertical Relational Distillation (VRD) and Cross Relational Distillation (XRD). VRD aligns teacher–student intra-modal similarity distributions to enforce consistent distillation strength across image and text embeddings. Meanwhile, XRD aligns the teacher-image–student-text and teacher-text–student-image similarity distributions to impose bidirectional cross-modal symmetry.
By jointly modeling these multidirectional relational structures, CLIP-RD aligns the student’s embedding geometry more faithfully to the teacher’s, outperforming CLIP-KD by $1.8\%p$. This performance improvement is maintained across diverse architectures, teacher scales, retrieval tasks, downstream tasks, and corruption settings, with negligible additional training-time overhead.
\end{abstract}
    
\begin{figure}[t]
    \centering
    \includegraphics[width=0.8\linewidth]{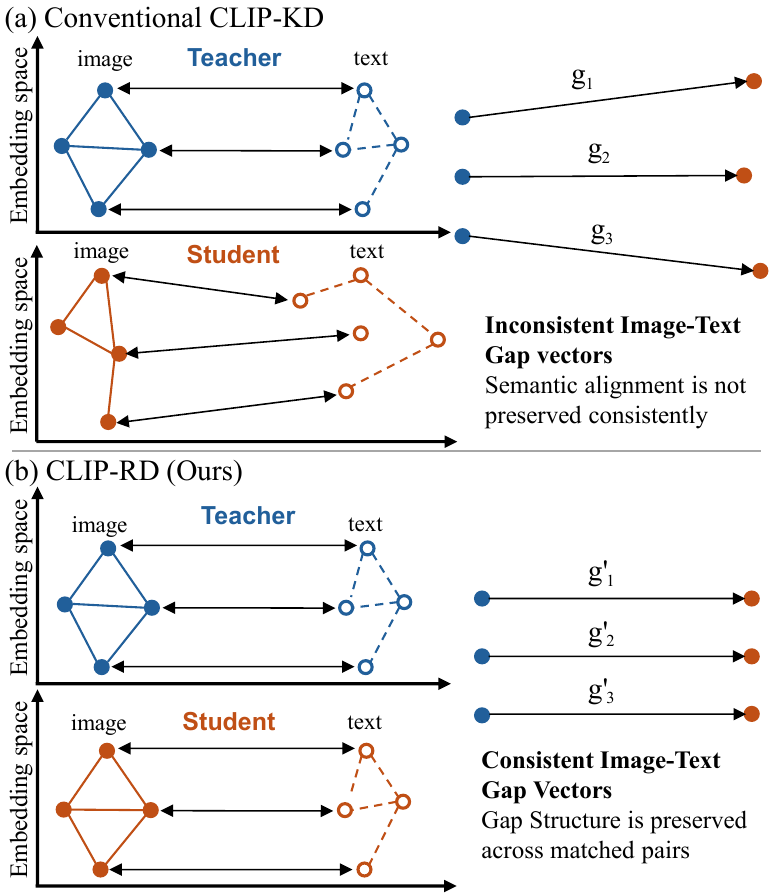}
    \caption{Comparison of embedding spaces between conventional CLIP-KD and the proposed CLIP-RD. \textbf{(a)} In CLIP-KD, the teacher and student exhibit less consistent image-text gap structures. \textbf{(b)} In contrast, our CLIP-RD better preserves the gap structure across matched image-text pairs, maintaining consistent gap vectors.}
    \label{fig:gap_vector}
    \vspace{-8pt}
\end{figure}

\section{Introduction}
\label{sec:intro}

CLIP~\cite{clip} is one of the most representative contrastive learning-based vision-language models, achieving strong generalization capability across diverse downstream tasks. This widespread adoption, however, comes with a significant computational cost due to its large-scale architecture, and knowledge distillation (KD)~\cite{KD} has thus gained considerable attention as an effective model compression technique~\cite{MGCLIP,PromtKD,DCLIP,TernaryCLIP}. However, distilling CLIP is particularly challenging, as it requires transferring both features and their relationships from teacher to student. Among prior works~\cite{TinyCLIP, CLIP-KD} that attempt this, CLIP-KD~\cite{CLIP-KD} proposes a framework that distills three types of knowledge, achieving state-of-the-art zero-shot accuracy on ImageNet.

\begin{figure*}[t]
    \centering
    \includegraphics[width=1.0\linewidth]{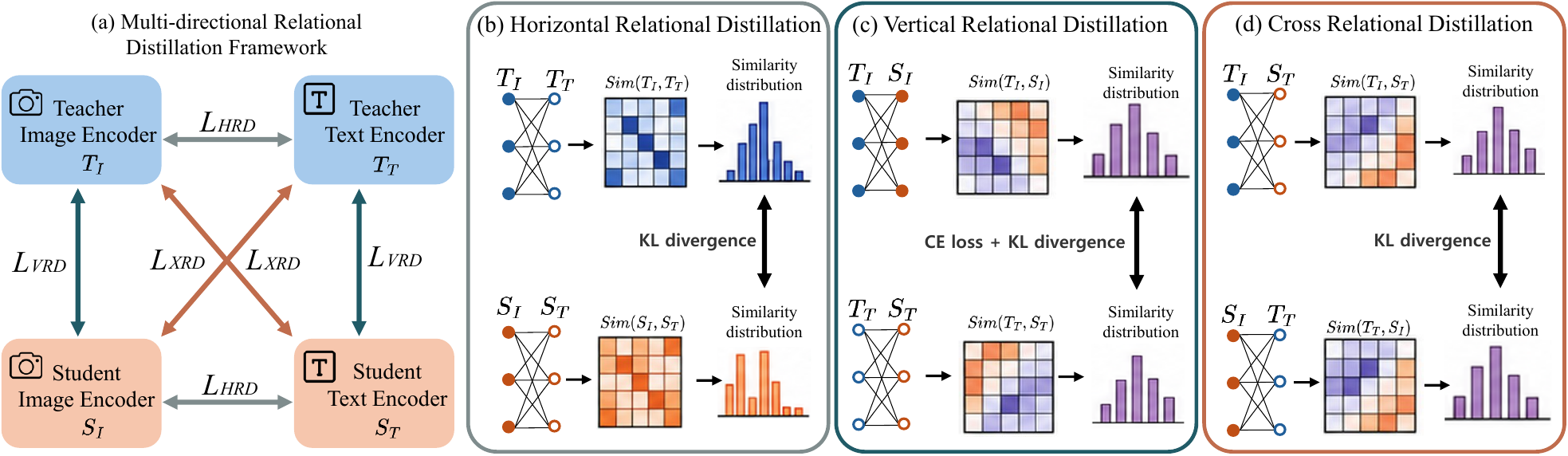}
    \caption{Illustration of the proposed CLIP-RD framework. \textbf{(a)} Our unified framework integrates these losses ($L_{HRD}$, $L_{VRD}$, $L_{XRD}$) to robustly preserve the complex geometric structure of the teacher's embedding space in the student model. \textbf{(b)} HRD for intra-model modality alignment, \textbf{(c)} VRD for inter-model uni-modal alignment, and \textbf{(d)} XRD for inter-model cross-modal alignment.}
    \label{fig:method}
    \vspace{-10pt}
\end{figure*}

Specifically, CLIP-KD~\cite{CLIP-KD} introduces three core distillation components: Feature Distillation (FD), Interactive Contrastive Learning (ICL), and Contrastive Relational Distillation (CRD). Among them, CRD aligns image-to-text similarity distributions within the model, enabling the student to learn the relational structure of the teacher's embeddings. As illustrated in Figure~\ref{fig:method}-(b), we refer to CRD as Horizontal Relational Distillation (HRD) throughout this paper since it aligns horizontal relational structures. However, it relies solely on unidirectional relational distillation, which limits effective alignment in the student’s representation space. To overcome this limitation, we propose CLIP-RD.

More specifically, HRD alone does not explicitly model several teacher-student relational structures, including vertical and cross-directed relationships, as shown in Figure ~\ref{fig:gap_vector}. Using orthogonal transformation, we show that HRD constrains only the horizontal geometric structure between image and text pairs within each model, leaving the vertical and cross-directed relationships under-constrained. Since an appropriately maintained modality gap is known to improve zero-shot performance~\cite{liang2022mind}, this unconstrained freedom may allow the gap structure to drift during distillation, potentially degrading zero-shot performance.This indicates that faithfully preserving the teacher's modality gap structure requires jointly modeling all three relational directions, rather than the horizontal direction alone. In later sections, we provide the full derivation of this orthogonal-transformation analysis, together with gap vector experiments showing that CLIP-RD's student embeddings exhibit a vertical and cross gap structure more closely aligned with the teacher's than CLIP-KD's.

To capture and transfer these geometric structures, we introduce Vertical Relational Distillation (VRD) and Cross Relational Distillation (XRD). This formulation explicitly models the multi-directional dependencies between the teacher and the student. VRD aligns vertically-directed similarity distributions using two losses, as shown in Figure~\ref{fig:method}-(c): contrastive loss and KL-divergence loss. We maximize the similarity between teacher and student embeddings via contrastive loss, and further align these distributions with KL-divergence loss, enforcing consistency in teacher-student distillation strength across modalities at the distribution level. XRD aligns cross-directed similarity distributions, as shown in Figure~\ref{fig:method}-(d), by constructing teacher text–student image and teacher image–student text pairs, and minimizing their KL-divergence. Through XRD, we enforce bidirectional symmetry in cross-modal teacher-student similarity distributions. 

With these two novel strategies together with HRD, we are able to model multi-directional relational distributions: horizontal, vertical, and cross, as illustrated in Figure~\ref{fig:method}-(a). Through this multi-directional framework, we effectively align the student's representation space and transfer CLIP's strong generalization capabilities to the student model. In our experiments, we use a CLIP model with a ViT-L/14~\cite{ViT} pre-trained on LAION-400M~\cite{LAION-400M}, then distill its knowledge into a lightweight ViT-B/16 student model using CC3M~\cite{CC3M} and CC12M~\cite{CC12M} datasets. We evaluate the student model on ImageNet~\cite{ImageNet} for zero-shot classification and on MSCOCO~\cite{MSCOCO} and Flickr30K~\cite{Flickr} for zero-shot retrieval. Our proposed method, CLIP-RD, demonstrates the effectiveness of our multi-directional relational distillation framework by outperforming CLIP-KD by 1.8\%p on ImageNet zero-shot classification, up to 6.7\%p in downstream tasks, up to 2.0\%p in distribution shift, and up to 1.0\%p in other architectures.

Our main contributions are as follows:

\begin{itemize}
    \item We identify that existing CLIP distillation methods rely on unidirectional relational distillation, which limits effective alignment in the student representation space. To address this limitation, we propose CLIP-RD, a unified multi-directional relational distillation framework.
    \item We propose Vertical Relational Distillation (VRD) and Cross Relational Distillation (XRD), which jointly model bidirectional symmetry across modalities and cross-modal teacher–student alignment by enforcing distribution-level consistency.
    \item Our method achieves $57.2\%$ zero-shot classification on ImageNet, outperforming CLIP-KD by $1.8\%p$, and demonstrates competitive performance on zero-shot retrieval benchmarks including MSCOCO and Flickr30K.
\end{itemize}
\section{Related Work}
\label{sec:related_work}
\subsection{Knowledge Distillation}
While scaling laws~\cite{SLNLM} have drastically improved model capabilities, the resulting latency and resource costs~\cite{ELLM} make model efficiency an active area of research~\cite{PQK,KLdivMSE,bert,tinybert,attention,TinyViT} across diverse tasks~\cite{MGCLIP,CVLM,CLIPTD} and modalities~\cite{attention,TinyViT,KDviaTA,bert,tinybert}. Knowledge Distillation (KD)~\cite{KD} has emerged as a key paradigm. KD transfers rich representations from a high-capacity teacher to a compact student, allowing the student to achieve superior performance despite the limitations of its smaller architecture. Although KD was initially established within the context of unimodal models~\cite{tinybert,TinyViT}, researchers have recently extended these methodologies to Vision-Language Models (VLMs)~\cite{MGCLIP,PromtKD,DCLIP,TernaryCLIP}. In this work, we address the limited exploration of relational structures in multimodal distillation by focusing on knowledge transfer that preserves cross-modal relationships.
\vspace{-5pt}

\subsection{CLIP and CLIP-KD}
CLIP~\cite{clip}, introduced by OpenAI, is a Vision-Language Model (VLM) trained through contrastive learning to learn rich representations. The core objective of CLIP is to optimize a symmetric contrastive alignment loss, which pulls the embeddings of paired image-text samples closer while pushing unpaired samples apart in a shared embedding space. This alignment enables CLIP to achieve strong zero-shot generalization~\cite{ZSGCLIP,ZSGR,clip,ZSTL,LiT} across various downstream tasks~\cite{FVLM,BEiTv2,MVP}, such as image classification~\cite{pmtVLM,AANLP,AlignCLIP} and cross-modal retrieval~\cite{CMRMI,AlignCLIP}, often outperforming existing VLMs~\cite{clip}.

As demand for efficient VLMs increases, KD has been widely used to compress large-scale pre-trained CLIP models~\cite{CLIP-KD,TinyCLIP,PromtKD,DCLIP,TernaryCLIP}. In this context, a notable advancement in compressing Vision-Language Models is CLIP-KD~\cite{CLIP-KD}, which introduces a comprehensive framework incorporating feature, relation, gradient, and contrastive paradigms. Rather than relying on a single objective, CLIP-KD empirically determines effective combinations of various distillation strategies. However, it still fails to fully capture relational distillation. Building upon this framework, we extend relational distillation by explicitly modeling vertical and cross-directed teacher-student relationships. This enables more comprehensive transfer of intra- and inter-modal relational structures from the teacher to the student.

\subsection{Gap Structure}
The modality gap was first introduced in \cite{liang2022mind}. It refers to the geometric phenomenon in which embeddings of different modalities (e.g., image and text) occupy distinct, separated regions within the shared representation space of contrastive vision-language models such as CLIP. \citet{liang2022mind} empirically demonstrates that adjusting the magnitude of the modality gap has a significant impact on downstream zero-shot classification performance and fairness, with an appropriately maintained gap leading to improved performance. To quantify this modality gap, \cite{liang2022mind} also defines the gap vector as the difference between the mean embedding vectors of the two modalities.

In this paper, we propose a new framework, CLIP-RD, which more faithfully distills the teacher's gap structure to the student.
\section{Methodology}

In this section, we present the CLIP-RD framework, comprising two novel distillation strategies, VRD and XRD. We denote $v_k^{T}$ and $v_k^{S}$ as the image embedding vectors from the teacher and student, respectively, and $s_k^{T}$ and $s_k^{S}$ as the corresponding text embeddings. Given a mini-batch $\mathcal{B} = \{({I}_k,{T}_k)\}\textstyle_{k=1}^{|\mathcal{B}|}$ where ${I}_k$ and ${T}_k$ denote the image and text of the $k$-th pair, respectively, we compute the relational distributions over these embedding vectors.

\subsection{Overview of CLIP}
Formally, the standard CLIP objective is formulated as a symmetric InfoNCE loss, consisting of image-to-text and text-to-image contrastive objectives. Each directional loss increases the similarity of matched image-text pairs relative to all unmatched pairs within the batch. The final CLIP loss, $\mathcal{L}_{task}=\mathcal{L}_{CLIP}$ is computed as the average of these two directional losses.

\subsection{Background of CLIP-KD}
To provide context and facilitate a better understanding, we briefly introduce the conventional method, CLIP-KD~\cite{CLIP-KD}. Detailed mathematical formulations of the loss functions are provided in the supplementary material.

\textbf{Feature Distillation (FD)}: Directly aligns feature embeddings between teacher and student within the same modality by minimizing the squared $L_2$ distance between their respective representations.

\textbf{Interactive Contrastive Learning (ICL)}: Following the InfoNCE-based formulation, we contrast student embeddings from one modality with teacher embeddings from another modality. By consistently using the teacher's representations to form the contrastive batch, the student effectively learns the relationships between modalities. The total ICL loss is computed as the average of the two cross-modal directions: the student image-to-teacher text loss and the student text-to-teacher image loss.

\textbf{Horizontal Relational Distillation (HRD)}: As shown in Figure~\ref{fig:method}, HRD transfers the horizontal relational structure to the student, by aligning the horizontally-directed similarity distributions. HRD enables the student to learn well-structured semantic relationships, thereby improving the quality of its feature representations.

For compact notation, let $i \in \{T,S\}$ denote the teacher or student
model, and $m \in \{v,s\}$ denote the anchor modality, with $\bar{m}$
representing the opposite modality, i.e., $\bar{v}=s$ and $\bar{s}=v$.
The corresponding cross-modal similarity distribution is defined as:

\begin{equation}
C_k^{i,(m \to \bar{m})}[j]
=
\frac{
\exp\left(
m_k^{i} \cdot \bar{m}_j^{i} / \tau^{i}
\right)
}{
\sum_{b=1}^{|\mathcal{B}|}
\exp\left(
m_k^{i} \cdot \bar{m}_b^{i} / \tau^{i}
\right)
},
\label{eq:hrd_dist}
\end{equation}

We denote the image-to-text and text-to-image distributions as
$p_k^i := C_k^{i,(v \to s)}$ and
$q_k^i := C_k^{i,(s \to v)}$, respectively.

\begin{equation}
\mathcal{L}_{\mathrm{HRD}, I \to T} =
\frac{1}{|\mathcal{B}|}
\sum_{k=1}^{|\mathcal{B}|}
\sum_{j=1}^{|\mathcal{B}|}
p_k^{T}[j]
\log
\frac{p_k^{T}[j]}{p_k^{S}[j]}
\end{equation}

\begin{equation}
\mathcal{L}_{\mathrm{HRD}, T \to I} =
\frac{1}{|\mathcal{B}|}
\sum_{k=1}^{|\mathcal{B}|}
\sum_{j=1}^{|\mathcal{B}|}
q_k^{T}[j]
\log
\frac{q_k^{T}[j]}{q_k^{S}[j]}
\end{equation}

\begin{equation}
\mathcal{L}_{\mathrm{HRD}} = 
\mathcal{L}_{\mathrm{HRD}, I \to T}
+
\mathcal{L}_{\mathrm{HRD}, T \to I}
\label{eq:L_HRD}
\end{equation}

Through KL divergence, the student effectively imitates the teacher’s horizontal relational structure across text and visual embeddings. This allows the student to mimic better structured semantic relations from the teacher, thereby improving the quality of feature representations.

\subsection{Vertical Relational Distillation (VRD)}
We introduce Vertical Relational Distillation (VRD), a distillation strategy that enforces consistency of teacher–student distillation strength across modalities at the distribution level. As illustrated in Figure~\ref{fig:method}-(c), VRD exploits two types of vertically-directed similarity distributions: image–image and text–text contrastive distributions. Specifically, we construct these distributions using both teacher and student as anchors. Given the image and text embeddings and learnable temperature parameters for each modality, each contrastive distribution is formulated as:

\begin{equation}
P_k^{m,(a \to b)}[j]
=
\frac{
\exp\left( m_k^{a} \cdot m_j^{b} / \tau^{m} \right)
}{
\sum_{c=1}^{|\mathcal{B}|}
\exp\left( m_k^{a} \cdot m_c^{b} / \tau^{m} \right)
}
\label{eq:vrd_dist}
\end{equation}

where $m \in \{v, s\}$ denotes the modality (image or text embedding), and $(a \to b)$ with $a, b \in \{T, S\}$, $a \neq b$, denotes the anchor direction. We write $I_k^{(a\to b)} := P_k^{v,(a\to b)}$ and $T_k^{(a\to b)} := P_k^{s,(a\to b)}$ for the image and text distributions, respectively, yielding $I_k^{TS}$, $I_k^{ST}$, $T_k^{TS}$, $T_k^{ST}$ in the following equations.

To effectively transfer these relational structures, VRD comprises two complementary losses: a contrastive loss and a KL-divergence loss. For the contrastive loss, we adopt an InfoNCE-based objective applied to each intra-modality distribution, maximizing the similarity between teacher and student embeddings of the same modality. Specifically, we compute this loss separately for each modality as $\mathcal{L}_\mathrm{CE-Image}$ and $\mathcal{L}_\mathrm{CE-Text}$. For $\mathcal{L}_\mathrm{CE-Image}$, we leverage two image–image similarity distributions, using the teacher and student as anchors respectively, and sum the resulting losses. $\mathcal{L}_\mathrm{CE-Text}$ follows the same formulation using text–text distributions. Finally, $\mathcal{L}_\mathrm{VRD-CE}$ is obtained by averaging $\mathcal{L}_\mathrm{CE-Image}$ and $\mathcal{L}_\mathrm{CE-Text}$. Given the intra-modality distributions, $\mathcal{L}_\mathrm{VRD-CE}$ is formulated as:
\begin{equation}
\mathcal{L}_{\mathrm{CE\text{-}Image}}
=
- \frac{1}{|B|} \sum_{k=1}^{|B|} ( \log I_k^{TS}[k]
+
\log I_k^{ST}[k] )
\end{equation}
\begin{equation}
\mathcal{L}_{\mathrm{CE\text{-}Text}}
=
- \frac{1}{|B|} \sum_{k=1}^{|B|} (  \log T_k^{TS}[k]
+
\log T_k^{ST}[k] )
\end{equation}
\begin{equation}
\mathcal{L}_{\mathrm{VRD\text{-}CE}}
=
\frac{1}{2}
\left(
\mathcal{L}_{\mathrm{CE\text{-}Image}}
+
\mathcal{L}_{\mathrm{CE\text{-}Text}}
\right)
\end{equation}

Subsequently, we apply KL-divergence loss to align the intra-modality similarity distributions. This loss aligns the image similarity distribution with the text similarity distribution. This encourages consistent distillation strength between the teacher and student across modalities. Specifically, we apply this loss to both teacher-anchored and student-anchored distributions and average them to obtain $\mathcal{L}_\mathrm{VRD-KL}$, which is formulated as:
\begin{equation}
\mathcal{L}_{\mathrm{VRD\_TS}}
=
\frac{1}{|\mathcal{B}|}
\sum_{k=1}^{|\mathcal{B}|}
\sum_{j=1}^{|\mathcal{B}|}
T_k^{TS}[j]
\log
\frac{
T_k^{TS}[j]
}{
I_k^{TS}[j]
}
\end{equation}
\begin{equation}
\mathcal{L}_{\mathrm{VRD\_ST}}
=
\frac{1}{|\mathcal{B}|}
\sum_{k=1}^{|\mathcal{B}|}
\sum_{j=1}^{|\mathcal{B}|}
T_k^{ST}[j]
\log
\frac{
T_k^{ST}[j]
}{
I_k^{ST}[j]
}
\end{equation}
\begin{equation}
\mathcal{L}_{\mathrm{VRD\text{-}KL}}
=
\frac{1}{2}
\left(
\mathcal{L}_{\mathrm{VRD\_TS}}
+
\mathcal{L}_{\mathrm{VRD\_ST}}
\right)
\end{equation}
The total VRD loss is defined as the sum of $\mathcal{L}_\mathrm{VRD\text{-}CE}$ and $\mathcal{L}_\mathrm{VRD\text{-}KL}$. By jointly optimizing these two losses, VRD effectively transfers the teacher model's representations and vertically-directed dependencies into the student. 

\begin{equation}
\mathcal{L}_{\mathrm{VRD}}
=
\mathcal{L}_{\mathrm{VRD\text{-}CE}}
+
\mathcal{L}_{\mathrm{VRD\text{-}KL}}
\label{eq:L_VRD}
\end{equation}

\subsection{Cross Relational Distillation (XRD)}
We further introduce a new distillation method, Cross Relational Distillation (XRD), to effectively transfer cross-modal relational knowledge. XRD aligns inter-modality similarity distributions (image–text and text–image), as illustrated in Figure~\ref{fig:method}-(d). Most conventional distillation methods focus on feature-level or intra-modality alignment. However, our method captures the bidirectional relationships between image and text representations of teacher and student. Specifically, XRD consists of four pairwise similarity distributions, $\mathcal{R}^{T_i \rightarrow S_t}$, $\mathcal{R}^{T_t \rightarrow S_i}$, $\mathcal{R}^{S_i \rightarrow T_t}$, and $\mathcal{R}^{S_t \rightarrow T_i}$. Given the image and text embeddings and learnable temperature parameters, each similarity distribution is formulated as: 

\begin{equation}
\mathcal{R}_k^{a_m \to b_n}[j]
=
\frac{
\exp\left( m_k^{a} \cdot n_j^{b} / \tau \right)
}{
\sum_{c=1}^{|\mathcal{B}|}
\exp\left(m_k^{a} \cdot n_c^{b} / \tau \right)
}
\end{equation}
where $m,n \in \{v, s\}$, $m \neq n$, denote the two modalities, and $a,b \in \{T,S\}$, $a \neq b$, define the model direction.

Each distribution represents the pairwise similarity between samples across the two modalities and two networks. The XRD loss consists of $\mathcal{L}_{T \rightarrow S}$ and $\mathcal{L}_{S \rightarrow T}$. $\mathcal{L}_{T \rightarrow S}$ and $\mathcal{L}_{S \rightarrow T}$ minimize the KL-divergence between teacher-to-student directed distributions, and student-to-teacher directed distributions, respectively. Unlike VRD, which aligns vertically-directed distributions, XRD aligns cross-modal dependencies, and therefore applies KL-divergence in both directions and averages the results to ensure symmetric alignment. Specifically, $\mathcal{L}_{T \rightarrow S}$ is obtained by computing KL-divergence between $\mathcal{R}_k^{T_i \rightarrow S_t}[j]$ and $\mathcal{R}_k^{T_t \rightarrow S_i}[j]$, and $\mathcal{L}_{S \rightarrow T}$ follows a similar formulation applied to student-to-teacher directed distributions. Through this strategy, the student learns how each modality of one network semantically interacts with the other modality of the other network, thereby enabling richer cross-modal alignment. The formulas for $\mathcal{L}_{T \rightarrow S}$ and $\mathcal{L}_{S \rightarrow T}$ are as follows:

\begin{equation}
\begin{split}
\mathcal{L}_{a \to b} =
\frac{1}{2} \bigg( & \frac{1}{\vert{}\mathcal{B}\vert{}} \sum_{k=1}^{\vert{}\mathcal{B}\vert{}} \sum_{j=1}^{\vert{}\mathcal{B}\vert{}} \mathcal{R}_k^{a_m \to b_n}[j] \log \frac{ \mathcal{R}_k^{a_m \to b_n}[j] }{ \mathcal{R}_k^{a_n \to b_m}[j] } \\
+ & \frac{1}{\vert{}\mathcal{B}\vert{}} \sum_{k=1}^{\vert{}\mathcal{B}\vert{}} \sum_{j=1}^{\vert{}\mathcal{B}\vert{}} \mathcal{R}_k^{a_n \to b_m}[j] \log \frac{ \mathcal{R}_k^{a_n \to b_m}[j] }{ \mathcal{R}_k^{a_m \to b_n}[j] } \bigg)
\end{split}
\end{equation}
where $m,n \in \{v, s\}$, $m \neq n$, denote the two modalities, and $a,b \in \{T,S\}$, $a \neq b$, define the model direction.

The overall XRD loss, $\mathcal{L}_{\mathrm{XRD}}$, is obtained by averaging $\mathcal{L}_{T \rightarrow S}$ and $\mathcal{L}_{S \rightarrow T}$, and is formulated as:
\begin{equation}
\mathcal{L}_{\mathrm{XRD}}
=
\frac{1}{2}(\mathcal{L}_{T \to S}
+
\mathcal{L}_{S \to T})
\label{eq:L_XRD}
\end{equation}

This symmetric formulation further ensures bidirectional consistency between teacher and student relational mapping across modalities. Thus, the student model can learn the cross-modal structural knowledge captured by the teacher to improve semantic alignment between image and text embeddings. 
Consequently, as illustrated in Figure~\ref{fig:method}-(a), HRD, VRD, and XRD together constitute a multi-directional relational distillation framework, enabling symmetric and balanced knowledge transfer from teacher to student. 

\subsection{CLIP-RD}
Our framework, CLIP-RD, is trained by minimizing a joint loss function: 
\begin{equation}
\mathcal{L}_{\mathrm{CLIP\text{-}RD}}
=
\mathcal{L}_{\mathrm{task}}
+
\alpha \mathcal{L}_{\mathrm{FD}}
+
\beta \mathcal{L}_{\mathrm{ICL}}
+
\mathbb{\lambda} \mathcal{L}_{\mathrm{RD}}
\end{equation}
where $\mathcal{L}_{\mathrm{RD}} =\mathcal{L}_{\mathrm{HRD}} + \mathcal{L}_{\mathrm{VRD}} + \mathcal{L}_{\mathrm{XRD}}$. $\alpha$, $\beta$, and $\mathbb{\lambda}$ are hyperparameters that balance the contributions of the different loss terms. In all experiments, we set $\alpha=2000, \beta=1$, and $\mathbb{\lambda}=1$ for all relational distillation losses, adopting the optimal scaling factors for $\mathcal{L}_{\mathrm{FD}}$, $\mathcal{L}_{\mathrm{ICL}}$, and $\mathcal{L}_{\mathrm{HRD}}$ reported in \cite{CLIP-KD}. This framework, $\mathcal{L}_{\mathrm{CLIP\text{-}RD}}$, delivers relational knowledge for more robust student models.

\subsection{Analysis}
The analysis below reveals the complementary roles of VRD and XRD: VRD preserves teacher–student relational information within each modality, while XRD constrains the mixed teacher–student cross-modal geometry that remains unconstrained by HRD.

\textbf{Why VRD and XRD Are Necessary}
HRD does not directly constrain cross-modal similarities involving mixed teacher-student representations. Consider a non-identity orthogonal transform $R$ such that $R^{\intercal}R=I$, and define transformed student embeddings as $\tilde{v}^S=Rv^S$ and $\tilde{s}^S=Rs^S$.
Then, 
\begin{equation}
(\tilde{v}^S)^{\intercal}\tilde{s}^S=(v^S)^{\intercal}R^{\intercal}Rs^S=(v^S)^{\intercal}s^S
\end{equation}
Thus, all within-student cross-modal similarities remain unchanged under a joint orthogonal transformation, making the HRD objective invariant to such transformations. However, the mixed teacher-student similarities $(v^T)^{\intercal}\tilde{s}^S = (v^T)^{\intercal}Rs^S$ and $(s^T)^{\intercal}\tilde{v}^S = (s^T)^{\intercal}Rv^S$ are generally not preserved. Therefore, HRD alone cannot uniquely constrain the orientation of the student embedding space relative to the teacher embedding space. XRD addresses this by directly and symmetrically aligning these mixed cross-modal distributions. Although ICL also uses mixed teacher-student pairs, it applies a contrastive objective to promote matched pairs over unmatched pairs rather than explicitly matching the full cross-directed similarity distributions. XRD complements ICL by enforcing distribution-level consistency between these relations. FD provides point-wise alignment but does not explicitly preserve relational structure across samples. VRD addresses this by aligning teacher–student similarity distributions within each modality, thereby explicitly constraining their intra-modal relational structure.

\textbf{Information-Theoretic Interpretation of VRD}
More specifically, VRD preserves teacher–student relational information within each modality. We show that its bidirectional InfoNCE objectives can be interpreted as maximizing a lower bound on the mutual information between teacher and student representations. The relevant proof is in the supplementary section.

\begin{table}[t]
\centering
\small
\caption{Performance comparison on zero-shot ImageNet-1K classification and cross-modal retrieval (ViT-L/14 teacher pre-trained on LAION-400M, ViT-B/16 student trained on CC3M+CC12M).}
\label{tab:main}
\renewcommand{\arraystretch}{1.3}
\begin{tabular}{lccccc}
\toprule
\multicolumn{1}{c}{\multirow{2}{*}{\textbf{Method}}} & \multirow{2}{*}{\textbf{IN-1K}} & \multicolumn{2}{c}{\textbf{MSCOCO}} & \multicolumn{2}{c}{\textbf{Flickr}} \\ \cline{3-4} \cline{5-6} 
\multicolumn{1}{c}{} & & \textbf{I2T} & \textbf{T2I} & \textbf{I2T} & \textbf{T2I} \\ \hline 
T: ViT-L/14 & 72.8 & 42.6 & 40.9 & 80.7 & 79.4 \\ \hline 
S: ViT-B/16 & 35.5 & 23.0 & 22.9 & 49.8 & 49.5 \\
CLIP-KD & 55.4 & 37.1 & 35.1 & 73.3 & 69.7 \\
\textbf{CLIP-RD (Ours)} & \textbf{57.2} & \textbf{37.8} & \textbf{36.7} & \textbf{73.7} & \textbf{71.7} \\ \bottomrule 
\end{tabular}
\vspace{-12pt}
\end{table}
\begin{table*}[t]
\vspace{-10pt}
\renewcommand{\arraystretch}{0.7}
\setlength{\tabcolsep}{2pt}
\centering
\caption{Comparison of zero-shot classification top-1 accuracy on diverse downstream tasks. Our CLIP-RD framework outperforms the conventional method across all datasets.}
\label{tab:zero_shot_results}
\begin{tabular}{lccccccc}
\toprule
\multicolumn{1}{c}{\textbf{Method}} & \textbf{CIFAR-10} & \textbf{CIFAR-100} & \textbf{Caltech101} & \textbf{EuroSAT} & \textbf{Food101} & \textbf{RESISC45} & \textbf{Sun397} \\ \midrule
T: ViT-L/14 & 94.7 & 77.7 & 88.4 & 41.3 & 84.9 & 63.2 & 71.5 \\ \midrule
S: ViT-B/16 & 79.5 & 39.4 & 75.8 & 19.1 & 36.2 & 33.0 & 45.6 \\
CLIP-KD     & 87.5 & 61.7 & 84.7 & 28.1 & 58.1 & 46.9 & 61.1 \\
\textbf{CLIP-RD (Ours)} & \textbf{87.9} & \textbf{62.7} & \textbf{85.3} & \textbf{34.8} & \textbf{60.0} & \textbf{48.7} & \textbf{62.7} \\ \bottomrule
\end{tabular}
\vspace{-10pt}
\end{table*}

\section{Experiments}
\subsection{Experimental Setup}
\label{sec:setup}
\subsubsection{Architecture}
We build CLIP-RD upon the original CLIP~\cite{clip} framework. For our primary evaluation, we distill knowledge from a ViT-L/14~\cite{ViT} teacher to a ViT-B/16 student; specifically, we utilize the pre-trained checkpoints provided by CLIP-KD~\cite{CLIP-KD} as our starting point for the teacher. To demonstrate the architecture-agnostic generalization of our method, we further conduct cross-architecture distillation using ViT-B/16 and ResNet-101~\cite{he2016deep} as teachers and diverse student architectures (MobileViT-S~\cite{MobileViT}, Swin-T~\cite{SwinTransformer}, EfficientNet-B0~\cite{EfficientNet}). The detailed configurations of visual and text encoders are presented in the supplementary material, following the OpenCLIP~\cite{openclip} codebase.

\subsubsection{Dataset}
The teacher model is initialized from CLIP-KD~\cite{CLIP-KD} checkpoints pre-trained on LAION-400M~\cite{LAION-400M}. For distilling the vision-language student model, we use Conceptual Captions 3M (CC3M)~\cite{CC3M} and 12M (CC12M)~\cite{CC12M} following the official annotation files. Images are downloaded from the provided URLs, and samples with inaccessible images are filtered out. After filtering, CC3M contains 2,230,103 training samples and 10,669 validation samples, while CC12M consists of 7,215,613 training samples.

For zero-shot classification evaluation, we use the ImageNet~\cite{ImageNet} validation set along with its distribution-shifted variants: ImageNet-V2 (IN-V2)~\cite{IN-V2}, ImageNet-Rendition (IN-R)~\cite{IN-Rendition}, and ImageNet-Sketch (IN-S)~\cite{IN-Sketch}, reporting the top-1 accuracy. Furthermore, we evaluate robustness to synthetic corruptions using ImageNet-C (IN-C)~\cite{IN-C}, reporting the mean Corruption Error (mCE) across five categories. For zero-shot cross-modal retrieval, we evaluate on MSCOCO~\cite{MSCOCO} and Flickr30K~\cite{Flickr} following the standard Karpathy split~\cite{Karpathysplit}, reporting Recall@1 (R@1) for both Image-to-Text (I2T) and Text-to-Image (T2I). We also evaluate zero-shot performance on a diverse set of benchmark datasets, including CIFAR-10/100~\cite{Cifar}, EuroSAT~\cite{eurosat}, RESISC45~\cite{RESISC45}, Food101~\cite{Food101}, Sun397~\cite{sun397}, and Caltech101~\cite{caltech101}.

\subsubsection{Training details}
We train the models for 32 epochs with a 10K iteration warmup, followed by a cosine learning rate decay schedule. Optimization is performed using AdamW~\cite{AdamW}, following the configuration of CLIP-KD, with an initial learning rate of 0.001 and a weight decay of 0.1. The global batch size is set to 1,024 across 8 workers. All learnable temperature parameters are initialized to 0.07 following CLIP-KD and the original CLIP.

Experiments are conducted on NVIDIA RTX PRO 6000 Blackwell GPUs. CLIP-RD introduces negligible additional training-time overhead compared with CLIP-KD, as detailed in the supplementary material. For other implementation details not explicitly specified, we follow the standard training protocols of CLIP-KD.

\begin{table}[t]
\renewcommand{\arraystretch}{1.0}
\setlength{\tabcolsep}{6pt}
\centering
\small
\footnotesize
\caption{Comparison of top-1 accuracy on ImageNet-1K and its distribution-shifted variants (IN-V2, IN-R, IN-S). Our CLIP-RD framework exhibits strong robustness against domain shifts, consistently outperforming the conventional method across all datasets. The teacher is pre-trained on LAION-400M and the students are trained on CC3M+CC12M.}
\label{tab:imagenet_results}
\resizebox{\columnwidth}{!}{%
\begin{tabular}{lcccc}
\toprule
\multicolumn{1}{c}{\textbf{Method}} & \textbf{IN-1K} & \textbf{IN-V2} & \textbf{IN-R} & \textbf{IN-S} \\ \midrule
T: ViT-L/14 & 72.8 & 65.5 & 84.7 & 59.6 \\ \midrule
S: ViT-B/16 & 35.5 & 31.1 & 46.8 & 24.5 \\
CLIP-KD     & 55.4 & 48.3 & 69.8 & 43.5 \\
\textbf{CLIP-RD (Ours)} & \textbf{57.2} & \textbf{49.4} & \textbf{71.8} & \textbf{44.8} \\ \bottomrule
\end{tabular}
}
\vspace{-10pt}
\end{table}

\subsection{Distilling CLIP Models}
\subsubsection{Zero-shot Transfer Performance}

In Table~\ref{tab:main}, we evaluate CLIP-RD on ImageNet for zero-shot classification, and MSCOCO and Flickr for zero-shot cross-modal retrieval. CLIP-RD achieves $57.2\%$ accuracy and outperforms the baseline of ViT-B/16 by $21.7\%p$ and CLIP-KD by $1.8\%p$. We observe that CLIP-RD improves I2T retrieval R@1 by $0.7\%p$ on MSCOCO and $0.4\%p$ on Flickr over CLIP-KD. For T2I retrieval, our framework also outperforms CLIP-KD by $1.6\%p$ on MSCOCO and $2.0\%p$ on Flickr. These results demonstrate that our multi-directional relational distillation framework effectively maximizes the alignment of the student’s representation space, leading to enhanced ImageNet zero-shot classification performance and zero-shot cross-modal retrieval performance.

As shown in Table~\ref{tab:zero_shot_results}, we evaluate zero-shot image classification performance across diverse datasets to assess the generalization capability of our model. Specifically, we select 7 datasets that were originally used to evaluate zero-shot performance in CLIP. Our proposed CLIP-RD outperforms CLIP-KD across all datasets. On object classification benchmarks such as CIFAR-10/100 and Caltech101, it surpasses CLIP-KD by $0.4\%p$ to $1.0\%p$. Notably, on EuroSAT, a challenging satellite image dataset, CLIP-RD demonstrates a substantial improvement of $6.7\%p$. We observe similar robust improvements on Food101 ($+1.9\%p$), Sun397 ($+1.6\%p$), and RESISC45 ($+1.8\%p$), confirming the strong zero-shot capability of our model.

\subsubsection{Robustness to Domain Shifts}

To verify the robustness of our model against domain shifts, we additionally evaluate on ImageNet variant datasets. As shown in Table~\ref{tab:imagenet_results}, CLIP-RD consistently outperforms CLIP-KD across all datasets, yielding improvements ranging from 1.1\%p to 2.0\%p. These results demonstrate that our method effectively transfers CLIP's strong generalization capability from teacher to student, maintaining high performance even under out-of-distribution scenarios.
\begin{figure}[t]
    \centering
    \includegraphics[width=0.6\linewidth]{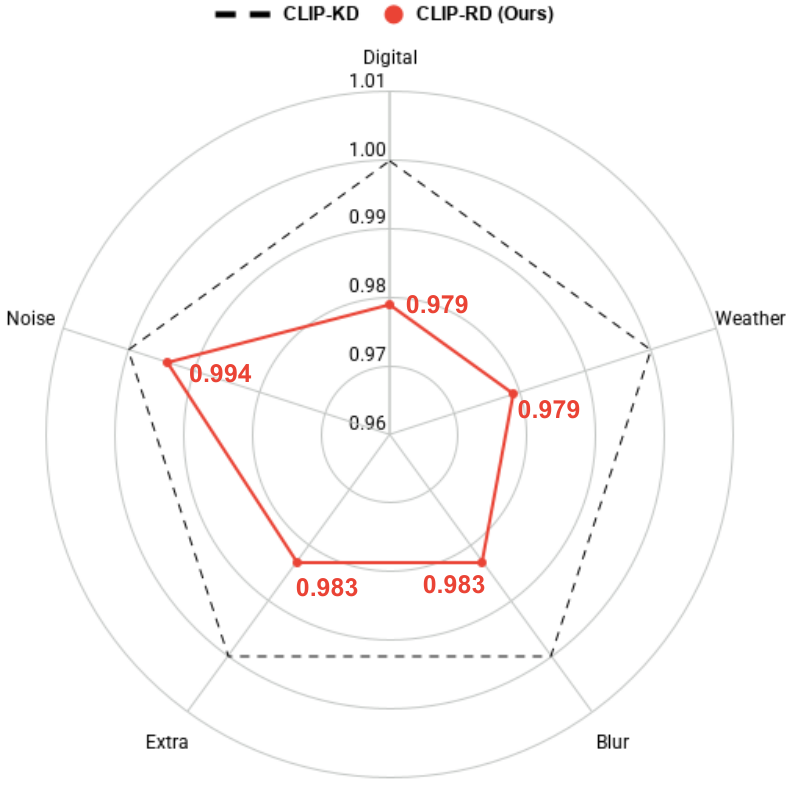}
    \caption{Relative mean Corruption Error (mCE) across five synthetic corruption categories and severity levels. Values are normalized against the CLIP-KD baseline (dashed line, $1.00$). Lower values indicate better robustness.}
    \label{fig:IN-C}
    \vspace{-10pt}
\end{figure}

\begin{table*}[t]
\vspace{-10pt}
\renewcommand{\arraystretch}{1.2}
\setlength{\tabcolsep}{3pt}
\centering
\caption{Zero-shot classification and robustness evaluation using various architectural combinations. Whether employing homogeneous or cross-architecture distillation (combining ViT and CNN families), CLIP-RD consistently demonstrates superior performance over the conventional method in all cases. All teachers are trained on CC3M and CC12M.}
\label{tab:architecture}
\resizebox{\textwidth}{!}{
\begin{tabular}{lcccc|lcccc}
\toprule
\multicolumn{1}{c}{\textbf{Method}} & \textbf{IN-1K}            & \textbf{IN-V2}         & \textbf{IN-R}          & \textbf{IN-S}          & \multicolumn{1}{c}{\textbf{Method}} & \textbf{IN-1K}            & \textbf{IN-V2}         & \textbf{IN-R}          & \textbf{IN-S}          \\ \hline
T: ViT-B/16                         & 35.5                   & 31.1                   & 46.8                   & 24.5                   & T: ResNet-101                       & 35.5                   & 31.0                   & 48.4                   & 25.6                   \\ \hline
S: MobileViT-S                      & 31.6                   & 26.4                   & 38.8                   & 19.3                   & S: MobileViT-S                      & 31.6                   & 26.4                   & 38.8                   & 19.3                   \\
CLIP-KD                             & 35.3                   & 30.0                   & 43.9                   & 22.8                   & CLIP-KD                             & 34.3                   & 29.3                   & 44.2                   & 22.9                   \\
\textbf{CLIP-RD (Ours)}             & \textbf{36.1 (+0.8 ↑)} & \textbf{30.9 (+0.9 ↑)} & \textbf{45.7 (+1.8 ↑)} & \textbf{23.6 (+0.8 ↑)} & \textbf{CLIP-RD (Ours)}             & \textbf{35.2 (+0.9 ↑)} & \textbf{30.2 (+0.9 ↑)} & \textbf{45.0 (+0.8 ↑)} & \textbf{23.7 (+0.8 ↑)} \\ \hline
S: Swin-T                           & 34.7                   & 29.7                   & 43.8                   & 23.4                   & S: Swin-T                           & 34.7                   & 29.7                   & 43.8                   & 23.4                   \\
CLIP-KD                             & 38.3                   & 32.9                   & 50.0                   & 26.9                   & CLIP-KD                             & 37.9                   & 32.3                   & 50.5                   & 27.4                   \\
\textbf{CLIP-RD (Ours)}             & \textbf{39.3 (+1.0 ↑)} & \textbf{34.5 (+1.6 ↑)} & \textbf{51.3 (+1.3 ↑)} & \textbf{27.9 (+1.0 ↑)} & \textbf{CLIP-RD (Ours)}             & \textbf{38.8 (+0.9 ↑)} & \textbf{33.4 (+1.1 ↑)} & \textbf{52.1 (+1.6 ↑)} & \textbf{28.1 (+0.7 ↑)} \\ \hline
S: EfficientNet-B0                  & 30.9                   & 26.4                   & 39.8                   & 19.6                   & S: EfficientNet-B0                  & 30.9                   & 26.4                   & 39.8                   & 19.6                   \\
CLIP-KD                             & 35.0                   & 30.7                   & 44.5                   & 23.2                   & CLIP-KD                             & 34.1                   & 29.0                   & 44.8                   & 23.3                   \\
\textbf{CLIP-RD (Ours)}             & \textbf{35.8 (+0.8 ↑)} & \textbf{30.8 (+0.1 ↑)} & \textbf{45.3 (+0.8 ↑)} & \textbf{23.6 (+0.4 ↑)} & \textbf{CLIP-RD (Ours)}             & \textbf{34.7 (+0.6 ↑)} & \textbf{29.5 (+0.5 ↑)} & \textbf{45.6 (+0.8 ↑)} & \textbf{23.9 (+0.6 ↑)} \\ \bottomrule
\end{tabular}
}
\vspace{-10pt}
\end{table*}

Furthermore, we evaluate the robustness against synthetic corruptions using ImageNet-C (IN-C)~\cite{IN-C}. As illustrated in Figure~\ref{fig:IN-C}, our CLIP-RD consistently reduces the relative mCE across diverse corruption categories compared to CLIP-KD. This indicates that our multi-directional relational distillation effectively reinforces the resilience of the student model against both domain shifts and synthetic corruptions.

\subsubsection{Architecture-Agnostic Distillation}
For this evaluation, we utilize a Transformer-based teacher (ViT) and a CNN-based teacher (ResNet) trained on the CC3M and CC12M datasets, and subsequently distill knowledge into diverse student models including MobileViT-S~\cite{MobileViT}, Swin-T~\cite{SwinTransformer}, and EfficientNet-B0~\cite{EfficientNet} using the same CC3M and CC12M training data. To demonstrate that our method is robust to architectural variations and model scales, we evaluate CLIP-RD on zero-shot image classification and domain-shift environments across these various teacher-student combinations.

As shown in Table~\ref{tab:architecture}, CLIP-RD consistently outperforms CLIP-KD across all evaluated teacher-student architecture combinations. These results demonstrate that the proposed relational objectives are applicable to both homogeneous and cross-architecture distillation settings.

In addition, we conducted additional experiments exploring a wider variety of dataset and model combinations. Specifically, we evaluated on a setting comprising a ViT-B/16 teacher (pre-trained on LAION-400M) and a student model utilizing either ViT-T/16 (distilled on CC3M and CC12M) or a ResNet backbone. Using these datasets, we measured performance on ImageNet classification and zero-shot cross-modal retrieval, including CC3M retrieval. Additional results and analyses for these experiments are provided in the supplementary material.

\begin{table}[t]
\centering
\caption{Ablation study demonstrating the effectiveness of Vertical Relational Distillation (VRD) and Cross Relational Distillation (XRD).}
\label{tab:ablation}
\resizebox{\linewidth}{!}{
\setlength{\tabcolsep}{8pt}
\begin{tabular}{lccccc}
\toprule
\multicolumn{1}{c}{\multirow{2}{*}{\textbf{Method}}} & \multirow{2}{*}{\textbf{IN-1K}} 
& \multicolumn{2}{c}{\textbf{MSCOCO}} 
& \multicolumn{2}{c}{\textbf{Flickr}} \\
\cline{3-6}
\addlinespace[2pt]
 &  & \textbf{I2T} & \textbf{T2I} & \textbf{I2T} & \textbf{T2I} \\
\addlinespace[-2pt]
\midrule
T: ViT-B/16     & 67.1        & 39.5         & 36.5         & 76.5         & 75.5        \\ \midrule
S: ViT-T/16     & 29.3        & 18.2         & 17.9         & 39.3         & 42.0        \\
CLIP-KD         & 41.3        & 27.4         & 24.1         & 58.4         & 56.4        \\
CLIP-KD+XRD     & 41.9        & 27.6         & 24.8         & 59.5         & 57.7        \\
CLIP-KD+VRD     & 42.0        & 27.6         & 25.0         & \textbf{61.0}         & \textbf{58.6}        \\
\textbf{CLIP-RD}         & \textbf{42.1}        & \textbf{27.8}         & \textbf{25.1}         & 58.3         & \textbf{58.6}       \\ \bottomrule
\end{tabular}
}
\vspace{-10pt}
\end{table}

\subsection{Ablation Study}
Table~\ref{tab:ablation} examines the effectiveness of Vertical Relational Distillation (VRD), Cross Relational Distillation (XRD), and the CLIP-RD approach with zero-shot ImageNet-1K classification and cross-modal retrieval (MSCOCO, Flickr) performance. For this experiment, we employ a ViT-B/16 pre-trained on LAION-400M as the teacher. The student is a ViT-T/16, which is trained from scratch on the CC3M~\cite{CC3M} and CC12M~\cite{CC12M} datasets, described in Section~\ref{sec:setup}.

First, both proposed components consistently outperform the baseline CLIP-KD ($41.3\%$ on IN-1K). Specifically, CLIP-KD+XRD and CLIP-KD+VRD bring improvements of $0.6\%p$ and $0.7\%p$ on IN-1K, along with enhancement across cross-modal retrieval tasks on MSCOCO and Flickr. By combining these components into our final framework, CLIP-RD achieves the highest IN-1K performance of $42.1\%$. Overall, both VRD and XRD provide consistent improvements over CLIP-KD across most evaluation metrics. Their joint use achieves the best performance on ImageNet and MSCOCO, suggesting that the two relational objectives provide complementary supervision.

\subsection{Gap Vector Experiment}
\begin{figure}
    \centering
    \includegraphics[width=0.8\linewidth]{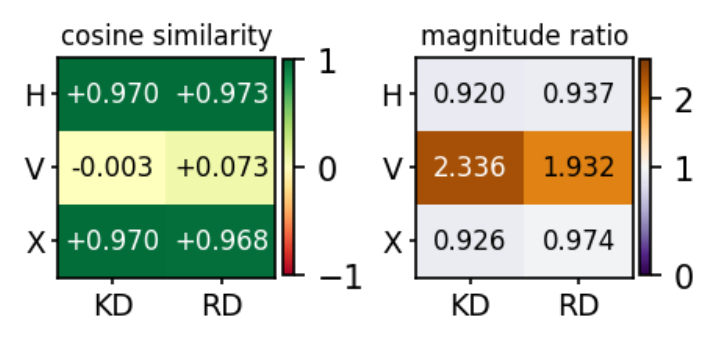}
    \caption{Gap vector comparison between CLIP-KD and CLIP-RD across horizontal ($H$), vertical ($V$), and cross ($X$) directions. We compare the corresponding gap vectors $g_1^d$ and $g_2^d$ using cosine similarity (left) and magnitude ratio (right).}
    \label{fig:heatmap}
    \vspace{-15pt}
\end{figure}

To examine how our method affects the teacher-student gap structure, we measure gap vectors along three directions and compare them with one another. \citet{liang2022mind} define only a horizontally directed gap vector. We therefore introduce two additional gap-vector directions, vertical and cross, to further characterize the teacher-student gap structure. Each gap vector is defined as follows, where v denotes the visual modality and s denotes the text modality.
\begin{align*}
\text{H}:& \{\text{mean}(v^T)\!-\!\text{mean}(s^T),\ \text{mean}(v^S)\!-\!\text{mean}(s^S)\},\\
\text{V}:& \{\text{mean}(v^T)\!-\!\text{mean}(v^S),\ \text{mean}(s^T)\!-\!\text{mean}(s^S)\},\\
\text{X}:& \{\text{mean}(v^T)\!-\!\text{mean}(s^S),\ \text{mean}(v^S)\!-\!\text{mean}(s^T)\}.
\end{align*}
For each direction $d\in\{H,V,X\}$, we denote the first and second vectors in the corresponding set as $g_1^d$ and $g_2^d$, respectively. We then compute the cosine similarity and magnitude ratio $(\|g_1^d\| / \|g_2^d\|)$, where a ratio closer to 1 indicates better geometric balance. As shown in Figure~\ref{fig:heatmap}, both CLIP-KD and CLIP-RD comparably preserve the horizontal gap direction, as both employ HRD. In the vertical gap direction, however, CLIP-KD yields a cosine similarity of -0.003, whereas CLIP-RD increases this to 0.073, and also achieves a magnitude ratio closer to 1 (1.932 vs. 2.336). In the cross direction, the two methods attain similar cosine similarity, while CLIP-RD achieves a magnitude ratio closer to 1 (0.974 vs. 0.926). These results suggest that CLIP-RD improves several aspects of the vertical and cross-directed gap geometry, particularly the balance of gap-vector magnitudes. Taken together, the results indicate that VRD and XRD provide additional constraints on the teacher-student gap structure beyond HRD alone, which may contribute to the improved zero-shot performance observed in our experiments.
\section{Conclusion}
In this paper, we proposed CLIP-RD, a multi-directional relational distillation framework that introduces two novel strategies: VRD and XRD. VRD aligns teacher-student similarity distributions within each modality using cross-entropy loss and KL-divergence, while XRD aligns the cross-modal distributions that mix teacher and student representations through KL-divergence. By jointly applying these complementary strategies, CLIP-RD provides additional constraints on the geometric correspondence between teacher and student embedding spaces. Extensive experiments demonstrate that incorporating these multi-directional relational structures improves multimodal knowledge distillation, outperforming CLIP-KD. We hope this work inspires future research to more effectively leverage the intrinsic relational characteristics of CLIP-like models.

\clearpage

\setcounter{table}{5}
\setcounter{figure}{4}
\setcounter{equation}{17}

\appendix
\section{Detailed Mathematical Formulations of CLIP-KD}
In this section, we introduce the primary loss functions employed in CLIP-KD~\cite{CLIP-KD}, a conventional knowledge distillation method.
\subsection{Feature Distillation (FD)}
CLIP-KD utilizes Feature Distillation (FD) to directly align feature embeddings between the teacher and student models within the same modality. Specifically, for a given batch $\mathcal{B}$, it minimizes the Mean Squared Error (MSE) between the visual embeddings ($v_k^{T}$ and $v_k^{S}$) and between the text embeddings ($s_k^{T}$ and $s_k^{S}$) generated by the teacher and student models. The feature distillation loss is formulated as:
\begin{equation}
\mathcal{L}_{\mathrm{FD}} =
\frac{1}{|\mathcal{B}|}
\sum_{k=1}^{|\mathcal{B}|}
\left(
\| v_k^{T} - v_k^{S} \|_2^2
+
\| s_k^{T} - s_k^{S} \|_2^2
\right).
\label{eq:L_FD}
\end{equation}

\subsection{Interactive Contrastive Learning (ICL)}
CLIP-KD utilizes Interactive Contrastive Learning (ICL) to minimize the contrastive loss. It enables the student to learn the relationships between modalities by contrasting student embeddings from one modality with teacher embeddings from another modality. The Interactive Contrastive Learning loss is formulated as: 
\begin{equation}
\mathcal{L}_{\mathrm{ICL}, I \to T}
=
- \log
\frac{
\exp\left( {v_k^{S} \cdot s_k^{T}/\tau} \right)
}{
\sum_{b=1}^{|\mathcal{B}|}
\exp\left( {v_k^{S} \cdot s_b^{T}/\tau} \right)
}.
\end{equation}

Similarly, the text-to-image objective is:

\begin{equation}
\mathcal{L}_{\mathrm{ICL}, T \to I}
=
- \log
\frac{
\exp\left( {s_k^{S} \cdot v_k^{T}/\tau} \right)
}{
\sum_{b=1}^{|\mathcal{B}|}
\exp\left( {s_k^{S} \cdot v_b^{T}/\tau} \right)
}.
\end{equation}

The total ICL loss is defined as the average of $\mathcal{L}_{\mathrm{ICL}, I \to T}$ and $\mathcal{L}_{\mathrm{ICL}, T \to I}$: 

\begin{equation}
\mathcal{L}_{\mathrm{ICL}} = \frac{1}{2}(\mathcal{L}_{\mathrm{ICL}, I \to T}+\mathcal{L}_{\mathrm{ICL}, T \to I})
\label{eq:L_ICL}
\end{equation}

\begin{table}[h]
\centering
\caption{Architecture details of the visual and text encoders.}
\label{tab:configuration}
\vspace{-5pt}
\resizebox{\columnwidth}{!}{
\begin{tabular}{llccccc}
\toprule
\multicolumn{3}{c}{\textbf{Visual Encoder}} & \multicolumn{4}{c}{\textbf{Text Encoder: Transformer~\cite{AAYN}}} \\
\cmidrule(lr){1-3} \cmidrule(lr){4-7}
\textbf{Model} & \textbf{Type} & \textbf{Params} & \textbf{Layer} & \textbf{Width} & \textbf{Head} & \textbf{Params} \\
\midrule
ViT-L/14~\cite{ViT} & \multirow{5}{*}{ViT} & 304.0M & 12 & 768 & 12 & 85.1M \\
ViT-B/16~\cite{ViT} & & 86.2M & 12 & 512 & 8 & 37.8M \\
\cmidrule{4-7}
ViT-T/16~\cite{ViT} & & 5.6M & \multirow{3}{*}{12} & \multirow{3}{*}{384} & \multirow{3}{*}{6} & \multirow{3}{*}{21.3M} \\
MobileViT-S~\cite{MobileViT} & & 5.3M & & & & \\
Swin-T~\cite{SwinTransformer} & & 27.9M & & & & \\
\midrule
ResNet-101~\cite{he2016deep} & \multirow{4}{*}{CNN} & 56.3M & \multirow{2}{*}{12} & \multirow{2}{*}{512} & \multirow{2}{*}{8} & \multirow{2}{*}{37.8M} \\
ResNet-50~\cite{he2016deep} & & 38.3M & & & & \\
\cmidrule{4-7}
ResNet-18~\cite{he2016deep} & & 11.4M & \multirow{2}{*}{12} & \multirow{2}{*}{384} & \multirow{2}{*}{6} & \multirow{2}{*}{21.3M} \\
EfficientNet-B0~\cite{EfficientNet} & & 4.7M & & & & \\ 
\bottomrule
\end{tabular}
}
\end{table}
\begin{table}[t]
\centering
\caption{Training time and computational overhead comparison between standard CLIP-KD and our proposed CLIP-RD.}
\label{tab:training_time}
\vspace{-5pt}
\resizebox{\columnwidth}{!}{
\begin{tabular}{llccc}
\toprule
\textbf{Architecture ($T \rightarrow S$)} & \textbf{Method} & \textbf{Training Time} & \textbf{$\Delta$ Time} & \textbf{Overhead} \\
\midrule
\multirow{2}{*}{ViT-L/14 $\rightarrow$ ViT-B/16} & CLIP-KD & 46h 11m & - & - \\
& \textbf{CLIP-RD (Ours)} & 45h 31m & -40m & -1.4\% \\
\midrule
\multirow{2}{*}{ViT-B/16 $\rightarrow$ ViT-T/16} & CLIP-KD & 27h 39m & - & - \\
& \textbf{CLIP-RD (Ours)} & 27h 53m & +14m & +0.8\% \\
\bottomrule
\end{tabular}
}
\vspace{-5pt}
\end{table}

\section{Experiment Details}
\subsection{Configuration}
Table~\ref{tab:configuration} presents the configurations of the image and text encoders used in our experiments. It includes the networks used in both the main experiments and the additional experiments described below. We use ViT-L/14, ViT-B/16, and ResNet-101 as teacher models, while the student models include ViT-B/16 alongside the other architectures, following the configuration details introduced in CLIP-KD~\cite{CLIP-KD} to ensure fair evaluation.

\subsection{Training Details}
Table~\ref{tab:training_time} presents the training time and computational overhead of our proposed CLIP-RD compared to the standard CLIP-KD baseline. Notably, the additional relational loss introduces negligible overhead to the overall training process. In the smaller-scale setup (ViT-B/16 $\rightarrow$ ViT-T/16), where both models are trained on the CC3M and CC12M datasets, CLIP-RD adds merely 14 minutes, corresponding to a marginal $+0.8\%$ overhead. Furthermore, we observe that as the model capacity and pretraining scale increase, this relative overhead diminishes. In the larger-scale setup (ViT-L/14 $\rightarrow$ ViT-B/16), which utilizes a teacher pretrained on LAION-400M and a student distilled on CC3M and CC12M, the training time actually decreases by 40 minutes ($-1.4\%$). Consequently, CLIP-RD is a highly efficient distillation framework that scales effectively to larger models and datasets without imposing additional training bottlenecks.

\begin{table}[t]
\centering
\caption{Comparison of zero-shot top-1 accuracy on ImageNet and its distribution shift variants across diverse student architectures.}
\label{tab:classification}
\setlength{\tabcolsep}{6pt}
\begin{tabular}{lcccc}
\hline
\textbf{Method}         & \textbf{IN-1K} & \textbf{IN-V2} & \textbf{IN-R} & \textbf{IN-S} \\ \hline
T: ViT-B/16             & 67.1           & 59.6           & 77.9          & 52.4          \\ \hline
S: ViT-T/16             & 29.3           & 24.9           & 34.2          & 16.9          \\
CLIP-KD                 & 41.3           & 35.5           & 46.3          & 26.3          \\
\textbf{CLIP-RD (Ours)} & \textbf{42.1}  & \textbf{36.2}  & \textbf{48.3} & \textbf{27.3} \\ \hline
S: ResNet-50            & 34.8           & 29.6           & 45.9          & 24.4          \\
CLIP-KD                 & 54.1           & 46.6           & 66.0          & 41.6          \\
\textbf{CLIP-RD (Ours)} & \textbf{55.1}  & \textbf{47.2}  & \textbf{66.9} & \textbf{42.3} \\ \hline
S: ResNet-18            & 27.3           & 22.8           & 35.0          & 18.0          \\
CLIP-KD                 & 38.2           & \textbf{31.9}  & 45.0          & 25.8          \\
\textbf{CLIP-RD (Ours)} & \textbf{38.6}  & \textbf{31.9}  & \textbf{46.2} & \textbf{26.2} \\ \hline
\end{tabular}
\end{table}
\begin{table}[t]
\centering
\caption{Comparison on zero-shot cross-modal retrieval performance, measured by Recall@1, on MSCOCO, Flickr, CC3M datasets across diverse student architectures.}
\label{tab:retrieval}
\setlength{\tabcolsep}{6pt}
\resizebox{\columnwidth}{!}{
\begin{tabular}{lcccccc}
\toprule
\multirow{2}{*}{\textbf{Method}} & \multicolumn{2}{c}{\textbf{MSCOCO}} & \multicolumn{2}{c}{\textbf{Flickr}} & \multicolumn{2}{c}{\textbf{CC3M}} \\
                                 & \textbf{I2T}     & \textbf{T2I}     & \textbf{I2T}     & \textbf{T2I}     & \textbf{I2T}    & \textbf{T2I}    \\ \hline
T: ViT-B/16                      & 39.5             & 36.5             & 76.5             & 75.5             & 43.8            & 42.2            \\ \hline
S: ViT-T/16                      & 18.2             & 17.9             & 39.3             & 42.0             & 33.6            & 34.0            \\
CLIP-KD                          & 27.4             & 24.1             & \textbf{58.4}    & 56.4             & 40.2            & 38.7            \\
\textbf{CLIP-RD (Ours)}          & \textbf{27.8}    & \textbf{25.1}    & 58.3             & \textbf{58.6}    & \textbf{40.6}   & \textbf{39.3}   \\ \hline
S: ResNet-50                     & 23.2             & 22.9             & 52.2             & 52.7             & 40.8            & 39.6            \\
CLIP-KD                          & 36.0             & \textbf{32.9}    & \textbf{74.9}    & 69.4             & 47.1            & 45.3            \\
\textbf{CLIP-RD (Ours)}          & \textbf{36.6}    & \textbf{32.9}    & 73.0             & \textbf{70.4}    & \textbf{47.2}   & \textbf{45.7}   \\ \hline
S: ResNet-18                     &16.2              &16.4               & 43.6             & 41.5            & 31.7            & 30.9            \\
CLIP-KD                          & 25.5             & 21.6             & \textbf{56.6}    & 52.4             & 36.9            & 34.1            \\
\textbf{CLIP-RD (Ours)}          & \textbf{25.6}    & \textbf{22.0}    & 54.4             & \textbf{52.7}    & \textbf{37.6}   & \textbf{34.5}   \\ \bottomrule
\end{tabular}
}
\end{table}

\section{Additional Experiments}
\subsection{Zero-shot Classification on ImageNet and its Variants}
Table \ref{tab:classification} presents the zero-shot classification top-1 accuracy on ImageNet (IN-1K)~\cite{ImageNet} and its distribution-shift variants, specifically ImageNet-V2 (IN-V2)~\cite{IN-V2}, ImageNet-Rendition (IN-R)~\cite{IN-Rendition}, and ImageNet-Sketch (IN-S)~\cite{IN-Sketch}. In this setup, we utilize a ViT-B/16~\cite{ViT} teacher pre-trained on LAION-400M~\cite{LAION-400M}, while the student models are distilled on the CC3M~\cite{CC3M} and CC12M~\cite{CC12M} datasets. To demonstrate the versatility of our approach, we perform knowledge distillation from the ViT teacher into both a homogeneous architecture (ViT-T/16) and heterogeneous CNN-based architectures (ResNet-50, ResNet-18)~\cite{he2016deep}. Notably, this includes transferring knowledge to highly compact and lightweight models. Across all settings, our proposed CLIP-RD consistently outperforms the standard CLIP-KD baseline. This confirms that relational knowledge effectively enhances zero-shot generalization and robustness against distribution shifts, demonstrating significant performance gains regardless of whether the student shares the teacher's architectural paradigm or has a fundamentally smaller capacity.

\subsection{Zero-shot Cross-Modal Retrieval on MSCOCO, Flickr, and CC3M}
Table \ref{tab:retrieval} presents the zero-shot cross-modal retrieval performance in terms of Recall@1, encompassing both Image-to-Text (I2T) and Text-to-Image (T2I) tasks, on the MSCOCO~\cite{MSCOCO}, Flickr~\cite{Flickr}, and CC3M~\cite{CC3M} datasets. Maintaining the exact experimental configuration used for classification, we utilize a LAION-400M-pre-trained ViT-B/16 as the teacher model and distill knowledge into various student architectures (ViT-T/16, ResNet-50, and ResNet-18) using the CC3M and CC12M datasets. The results demonstrate that our proposed CLIP-RD consistently outperforms the standard CLIP-KD baseline across the vast majority of retrieval scenarios. Notably, CLIP-RD yields consistent improvements in T2I retrieval across all datasets and student architectures. Whether transferring knowledge to a homogeneous lightweight Vision Transformer or heterogeneous CNN-based models, the injection of relational distillation ensures consistently robust cross-modal retrieval performance. This proves that our method better preserves the rich semantic alignment and the original geometric structure between images and texts compared to traditional distillation approaches, ensuring robust cross-modal understanding even in highly compact models.

\section{Analysis}
We analyze our framework in several ways using ViT-B/16 pre-trained on LAION-400M~\cite{LAION-400M} as the teacher model and ViT-T/16 distilled using CC3M~\cite{CC3M} and CC12M~\cite{CC12M} datasets as the student model. The following analyses provide a deeper understanding of the proposed method.
\subsection{Loss Analysis}
\begin{figure*}[h!]
\centering
\includegraphics[width=0.6\textwidth]{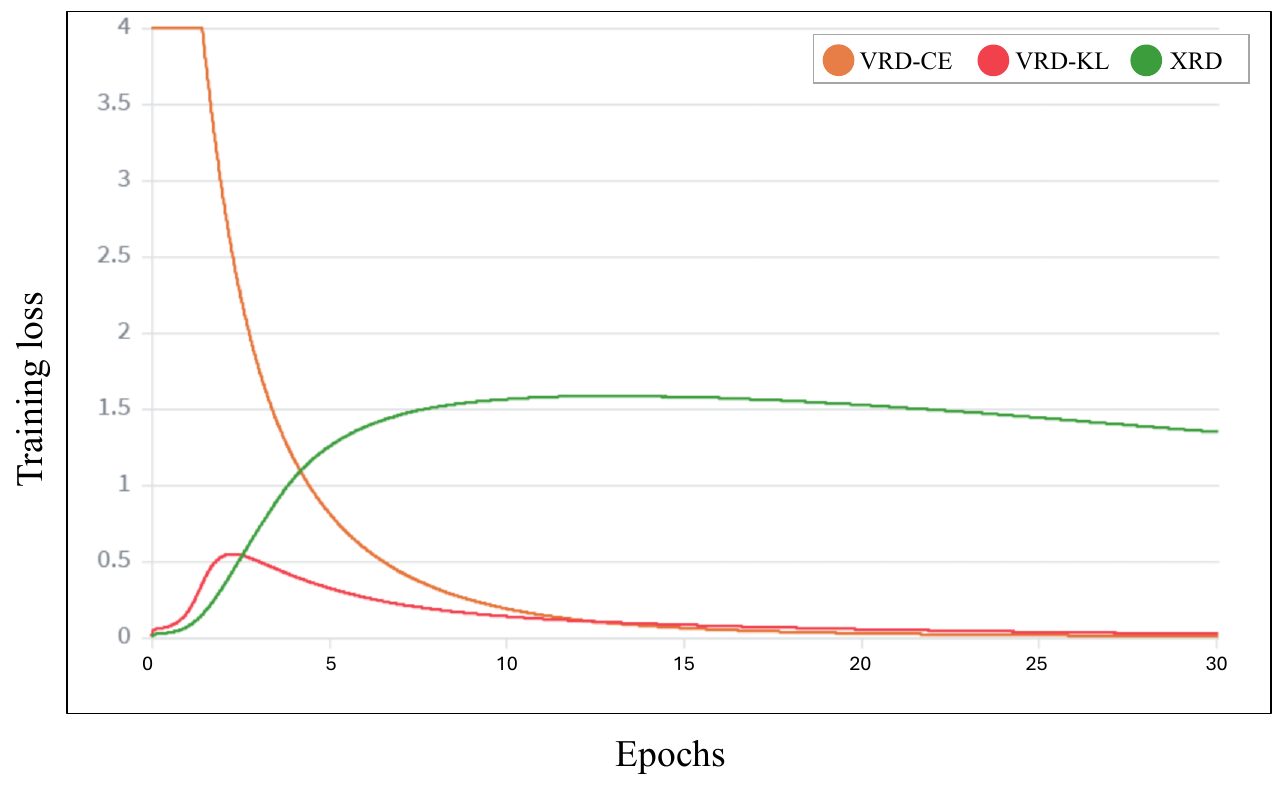}
\caption{Training loss. The y-axis is clipped at 4.0 to emphasize relative trends after the initial warm-up phase.}
\label{train_loss}
\end{figure*}
We conduct a loss analysis by measuring the training losses over epochs, with results shown in Figure~\ref{train_loss}. As illustrated, $\mathcal{L}_{\mathrm{VRD\text{-}CE}}$, $\mathcal{L}_{\mathrm{VRD\text{-}KL}}$, and $\mathcal{L}_{\mathrm{XRD}}$ all decrease as training progresses. Notably, although $\mathcal{L}_{\mathrm{VRD\text{-}KL}}$ and $\mathcal{L}_{\mathrm{XRD}}$ initially increase from very small values, they quickly stabilize and subsequently decrease. This indicates that our proposed strategies train the model in a stable and consistent manner.

\subsection{Training Curves Analysis}
\begin{figure*}[h!]
\centering
\includegraphics[width=0.8\textwidth]{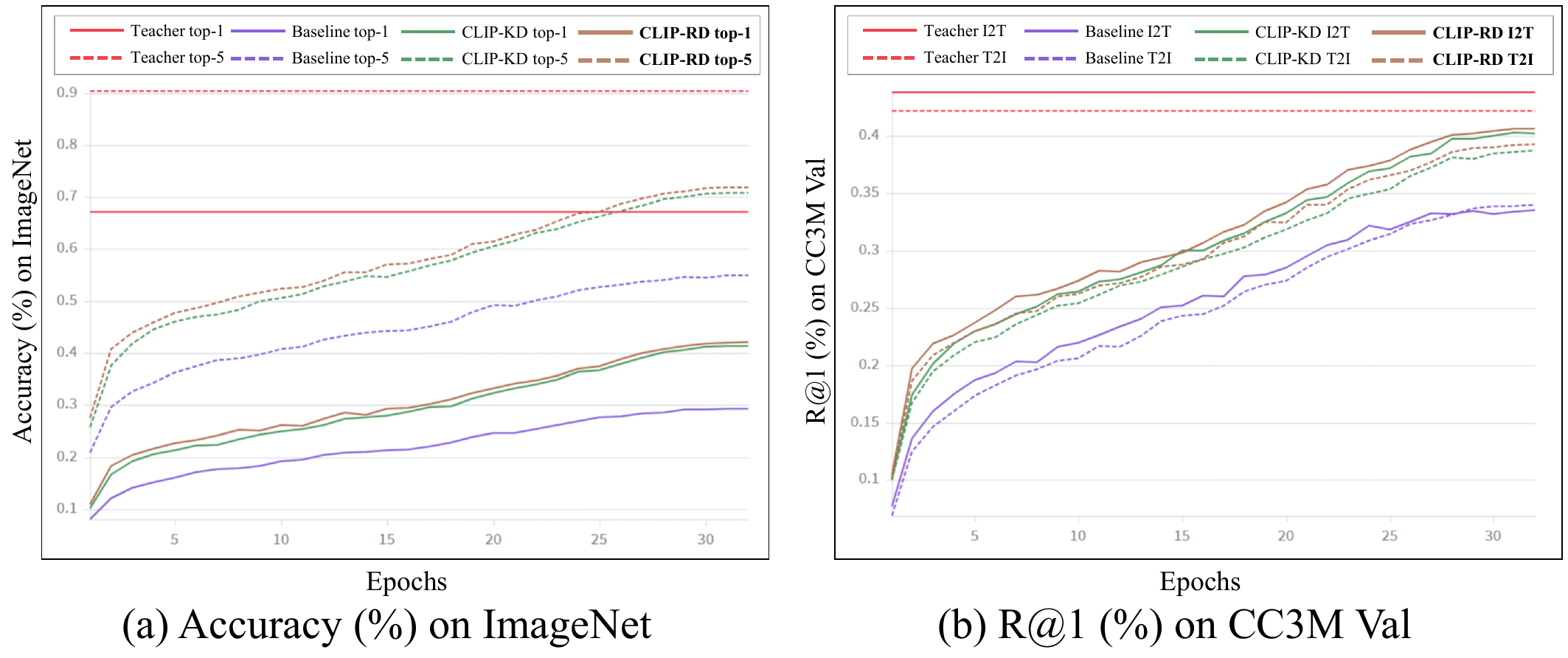}
\caption{Accuracy on IN and R@1 on CC3M Val.}
\label{in_rt}
\end{figure*}

We analyze the training performance curves of ImageNet top-1/top-5 accuracy and CC3M I2T/T2I R@1. As illustrated in Figure~\ref{in_rt}, CLIP-RD consistently achieves higher performance than CLIP-KD on both ImageNet and CC3M across all metrics and anchors. These results demonstrate the stability and robustness of our method throughout the training process.

\subsection{Similarity Distribution Analysis}
\begin{figure*}[h!]
\centering
\includegraphics[width=0.8\textwidth]{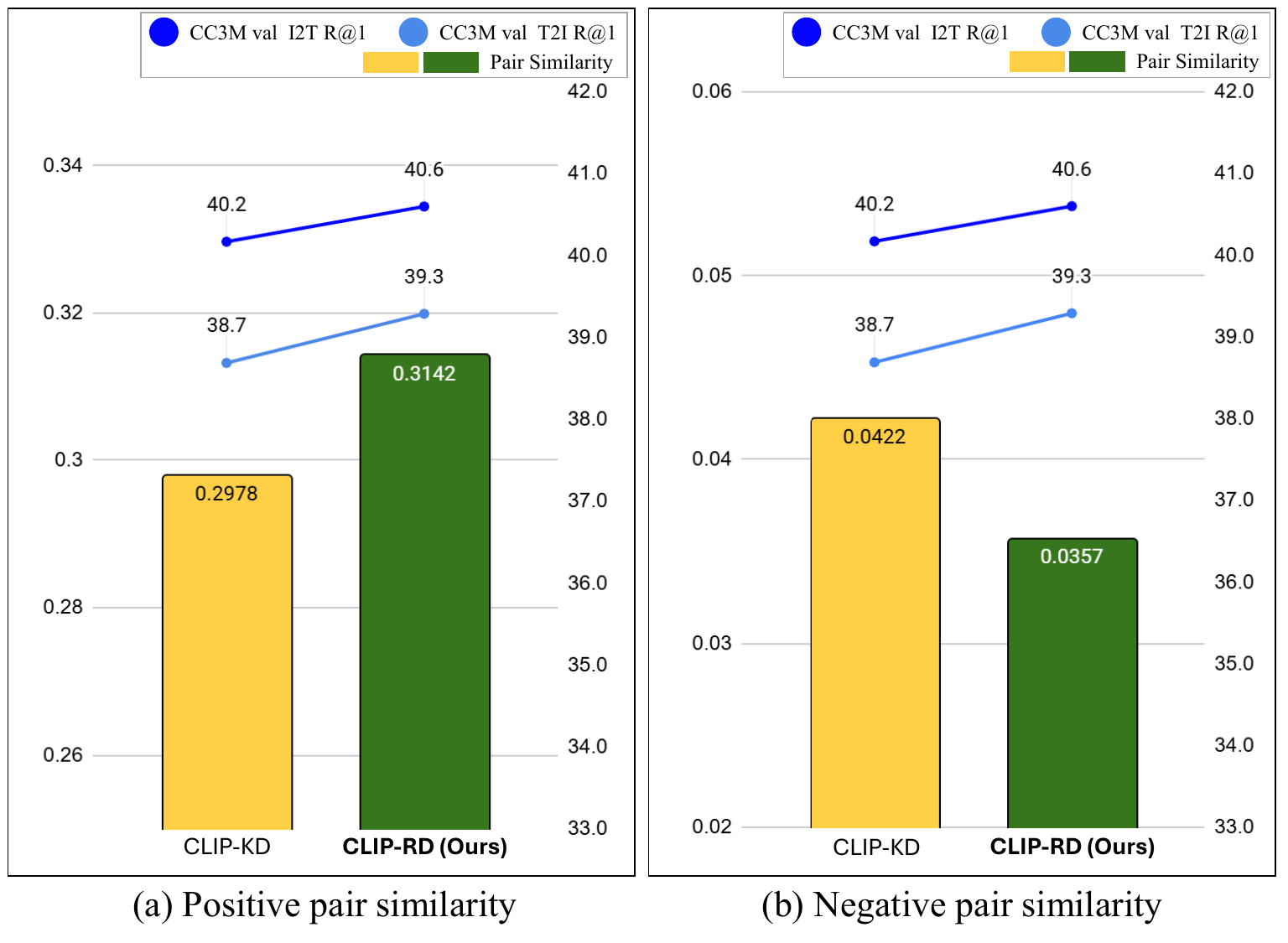}
\caption{Positive and negative pair similarity and CC3M validation set retrieval performance with CLIP-KD and CLIP-RD.}
\label{pair_sim}
\end{figure*}

\begin{figure*}[h!]
\centering
\includegraphics[width=0.8\textwidth]{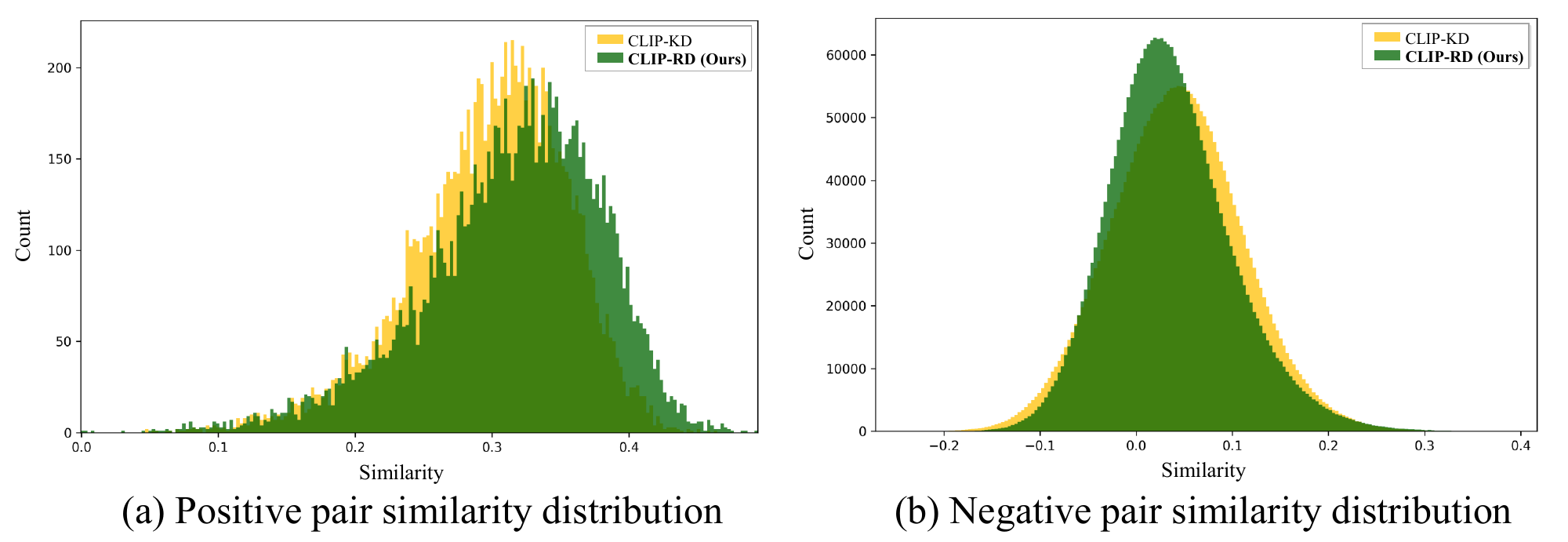}
\caption{Positive and negative pair similarities distribution with CLIP-KD and CLIP-RD. As in Figure~\ref{pair_sim}, higher positive (\textit{a}) and lower negative (\textit{b}) similarity indicate better alignment for representation space.}
\label{sim_dis}
\end{figure*}
We compare positive and negative pair similarities between CLIP-KD and our method on the CC3M validation subset. Specifically, we compute the average cosine similarity across all positive pairs and all negative pairs, respectively. Higher positive pair similarity and lower negative pair similarity indicate a more discriminative representation space with clearer boundaries between positive and negative pairs.

As illustrated in Figure~\ref{pair_sim}, CLIP-RD achieves higher positive pair similarity and lower negative pair similarity than CLIP-KD, indicating that our method pulls positive pairs closer while pushing negative pairs further apart. This is also consistent with the CC3M retrieval results.

We further compare the distributions of positive and negative pair similarities in Figure~\ref{sim_dis}. The positive pair similarity distribution shifts toward higher values compared to CLIP-KD, while the negative pair similarity distribution shifts toward lower values, demonstrating that our method effectively maximizes the alignment of the student's representation space.

\begin{figure*}[h!]
\centering
\includegraphics[width=0.6\textwidth]{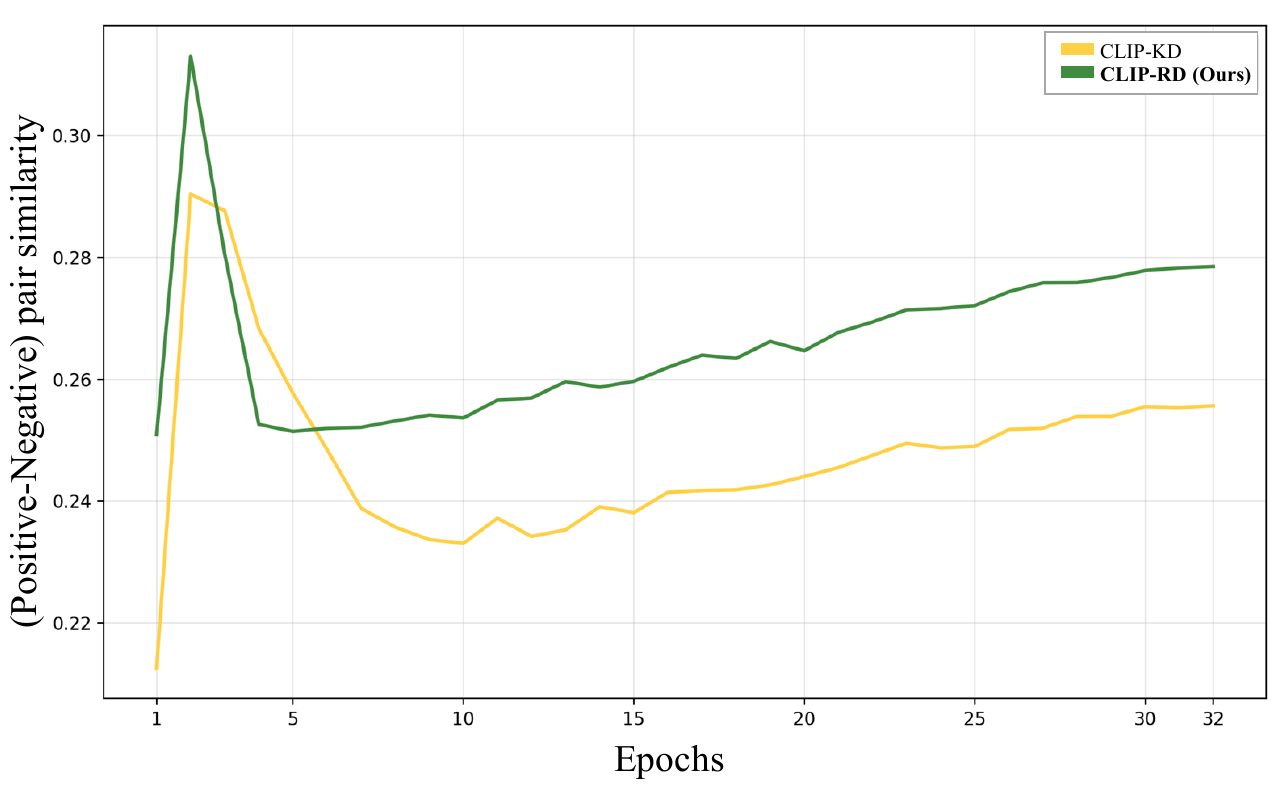}
\caption{(Pos-neg) pair similarity.}
\label{pos-neg}
\end{figure*}

Finally, we compare the positive-negative similarity gap between CLIP-KD and our method, defined as the difference between the mean positive and mean negative similarities over training epochs. This metric reflects the overall level of discriminability between positive and negative pairs. As shown in Figure~\ref{pos-neg}, CLIP-RD consistently outperforms CLIP-KD across most epochs, indicating that our method maintains superior alignment throughout training.

\subsection{Lower Bound of Mutual Information: Formulaic Insight of VRD-CE Loss}
Let $v^S \sim \mu(v^S), v^T \sim \mu(v^T)$ denote the student and teacher image embeddings. The joint distribution of positive pairs is $\mu(v^S, v^T)$ while negative pairs are sampled from the product of marginals $\mu(v^S)\mu(v^T)$. We introduce a binary variable $C \in \{0,1\}$ where $C=1$: positive pair, $C=0$: negative pair.

\begin{align}
    \begin{split}
        \eta(v^S,v^T|C=1) &= \mu(v^S,v^T)
        \\
        \eta(v^S,v^T|C=0) &= \mu(v^S)\mu(v^T)
    \end{split}
\end{align}

For each positive pair, there are $N$ negative pairs. The prior probabilities of $C$ are:
\begin{equation}
    \eta(C=1)=\frac{1}{1+N}, ~~\eta(C=0)=\frac{N}{1+N}
\end{equation}
for $N = \mathcal{B}-1$.

By Bayes' theorem,
\begin{align}
    \begin{split}
        &\eta(C=1|v^S,v^T) \\
        &\quad = \frac{\eta(v^S,v^T|C=1)\eta(C=1)}{\eta(v^S,v^T|C=1)\eta(C=1)+\eta(v^S,v^T|C=0)\eta(C=0)}\\
        &\quad = \frac{\mu(v^S,v^T)}{\mu(v^S,v^T)+N\mu(v^S)\mu(v^T)}
    \end{split}
\end{align}

The log-posterior can be written as:
\begin{align}
    \begin{split}
        \log{\eta(C=1|v^S,v^T)} 
        &= \log{\frac{\mu(v^S,v^T)}{\mu(v^S,v^T)+N\mu(v^S)\mu(v^T)}} \\ 
        &= -\log\left(1+N\frac{\mu(v^S)\mu(v^T)}{\mu(v^S,v^T)}\right) \\ 
        &\leq -\log{N} + \log\frac{\mu(v^S,v^T)}{\mu(v^S)\mu(v^T)}
    \end{split}
\end{align}

The expected log class posterior is related to the mutual information as follows:
\begin{align}\label{eq:stot}
    \begin{split}
        &\mathbb{E}_{\eta(v^S, v^T | C=1)} \log \eta(C = 1 | v^S, v^T) \\
        &\quad \leq -\log{N} + \mathbb{E}_{\mu(v^S,v^T)}\log\frac{\mu(v^S,v^T)}{\mu(v^S)\mu(v^T)} \\ 
        &\quad = -\log{N} + I(v^S, v^T) \\ 
        &\Leftrightarrow I(v^S, v^T) \geq \log{N} - \mathcal{L}_{\text{infoNCE}}(S \rightarrow T)
    \end{split}
\end{align}
where $I(v^S, v^T)$ denotes the mutual information between $v^S$ and $v^T$. Similarly,
\begin{align}\label{eq:ttos}
    \begin{split}
        &\mathbb{E}_{\eta(v^T, v^S | C=1)} \log \eta(C = 1 | v^T, v^S) \\
        &\quad \leq -\log{N} + I(v^T, v^S) \\ 
        &\Leftrightarrow I(v^T, v^S) \geq \log{N} - \mathcal{L}_{\text{infoNCE}}(T \rightarrow S)
    \end{split}
\end{align}

By combining Eq.~\ref{eq:stot} and Eq.~\ref{eq:ttos}:
\begin{align}
    \begin{split}
        &2I(v^S; v^T) \geq 2\log{N} - \mathcal{L}_{\text{infoNCE-Image}}(T \rightarrow S) \\
        &\quad - \mathcal{L}_{\text{infoNCE-Image}}(S \rightarrow T)
    \end{split}
\end{align}
The two directional InfoNCE-Image objectives provide lower bounds on the same mutual information. By combining the two directions, the symmetric objective maximizes a lower bound on the mutual information between teacher and student image representations.

Similarly,
\begin{align}
    \begin{split}
        &2I(s^S; s^T) \geq 2\log{N} - \mathcal{L}_{\text{infoNCE-Text}}(T \rightarrow S) \\
        &\quad - \mathcal{L}_{\text{infoNCE-Text}}(S \rightarrow T)
    \end{split}
\end{align}
Each of the two directional InfoNCE-Text objectives provides a lower bound on the same mutual information. By jointly optimizing both directions, the symmetric objective maximizes a lower bound on the text mutual information between the teacher and student text representations.

According to the InfoNCE bounds, the expected loss provides a lower bound on mutual information between teacher and student representations. Therefore, minimizing $\mathcal{L}_{\mathrm{VRD\text{-}CE}}$ approximately maximizes a lower bound on the mutual information.

\newpage
{
    \small
    \bibliographystyle{ieeenat_fullname}
    \bibliography{main}
}

\end{document}